\documentclass[11pt]{article}

\usepackage[final]{acl}

\usepackage{times}
\usepackage{latexsym}
\usepackage{xurl} % allows line breaks in URLs
\usepackage{hyperref}
\usepackage[most]{tcolorbox}
\usepackage{booktabs} % For nice rules
\usepackage{array}    % For better column formatting
\usepackage{tabularx} 
\usepackage{comment}

\usepackage[T1]{fontenc}
\usepackage[utf8]{inputenc}

\usepackage{microtype}

\usepackage{inconsolata}

\usepackage{graphicx}

\title{JGU Mainz's Submission to the WMT25 Shared Task on LLMs with Limited
Resources for Slavic Languages: MT and QA}

\author{
  \textbf{Hossain Shaikh Saadi,\textsuperscript{1}}
  \textbf{Minh Duc Bui,\textsuperscript{1}}\\
  \textbf{Mario Sanz-Guerrero,\textsuperscript{1}}   and 
  \textbf{Katharina von der Wense\textsuperscript{1,2}}\\
  \textsuperscript{1}Johannes Gutenberg University Mainz, Germany\\
  \textsuperscript{2}University of Colorado Boulder, USA
}

\begin{document}
\maketitle

\begin{abstract}
This paper presents the JGU Mainz submission to the WMT25 Shared Task on LLMs with Limited Resources for Slavic Languages: Machine Translation and Question Answering, focusing on Ukrainian, Upper Sorbian, and Lower Sorbian. For each language, we jointly finetune a Qwen2.5-3B-Instruct model for both tasks with parameter-efficient finetuning. Our pipeline integrates additional translation and multiple-choice question answering (QA) data. For Ukrainian QA, we further use retrieval-augmented generation. We also apply ensembling for QA in Upper and Lower Sorbian. Experiments show that our models outperform the baseline on both tasks.
\end{abstract}

\section{Introduction}
While large language models (LLMs) are strong multitask learners for high-resource languages such as English, this is not the case for smaller LLMs and languages with limited data. In this setting, a trade-off presents between the performance on different tasks.
%It is highly challenging to develop or train large language models to perform multiple tasks effectively in low-resource languages and even more difficult under computing resource constraints. 
The \textit{WMT25 Shared Task on LLMs with Limited Resources for Slavic Languages: MT and QA}\footnote{\url{https://www2.statmt.org/wmt25/limited-resources-slavic-llm.html}} focuses on the development of relatively small LLMs ($\le$3B parameters) that are capable of performing both machine translation (MT) and multiple-choice question answering (QA) in Slavic languages with limited amounts of data. Three languages -- a mid-resource language, Ukrainian (UK), and two severely low-resource languages, Upper Sorbian (HSB) and Lower Sorbian (DSB) -- are targeted, and only Qwen2.5 models with 0.5B, 1.5B, or 3B parameters are permitted. The MT source languages are Czech (CS) and English (EN) for Ukrainian as well as German (DE) for DSB and HSB.
%The primary aim is to encourage research on effective multilingual LLMs generalizes across tasks and languages under low-resource settings. 
%MT and QA for low-resource languages are challenging, due to the lack of publicly available datasets for the tasks. For the minimal-resource languages DSB and HSB, the issue is magnified further due to the scarcity of even monolingual data. 

Our proposed approach 
%addresses these challenges by building a multi-stage, parameter-efficient finetuning strategy using a 
consists of a Qwen2.5-3B-Instruct model \cite{qwen_report}, which is inherently multilingual
%. However, performance on these languages and tasks under consideration is significantly poor. We 
and which we jointly finetune on both MT and QA data, combining the provided resources with additional datasets we curate ourselves. For DSB and HSB MT, we enhance the training data using synthetic data generated through back-translation. We further add an additional parallel dataset for DSB. For QA, we add a total of 16 high-quality English MCQ datasets. All QA datasets are enhanced via automatic translation such that they are bilingual (English and the target language). For Ukrainian QA, we incorporate retrieval-augmented generation (RAG) using domain-relevant Wikipedia pages and 10 books related to the subjects of the provided Ukrainian MCQ dataset. 

At inference time, we use similarity-based few-shot in-context learning (ICL) for MT. For QA, we permute the order of the answer options, and average the probabilities for all options, to increase robustness against answer ordering biases \cite{positional-bias}. % remove cite if not needed 

%Our unified (MT and QA) LoRA \cite{lora} finetuning strategy is designed to maximize effectiveness within the 3-billion parameter limit. Furthermore, it allows us to assess the viability of a multilingual model in multiple tasks at the same time with limited resources and under parameter constraints. 
While not winning for any task--language combination, our primary submission consistently outperforms the baseline on all tasks, demonstrating the effectiveness of our approach. For DE--DSB translation, ChrF++ improves by over 55 points, while DE--HSB translation sees gains of over 65 points, reflecting substantial quality improvements. On DSB QA, accuracy increases by up to 12.34 percentage points, and, on HSB QA, accuracy increases by 10.27 points. For CS--UK and EN--UK MT, ChrF++ improves by 4.61 and 2.7, respectively, while, on Ukrainian QA, our submission outperforms the baseline by 4.66 accuracy points.

\section{Data}

\subsection{Provided Data}
\paragraph{Ukrainian} For EN--UK and CS--UK MT, no training data are provided. Only development sets are given, containing 6,263 CS--UK and 5,108 EN--UK parallel sentences. 

The UK QA data are curated from the External Independent Evaluation (3HO/ZNO), an exam for admission to Ukrainian universities. The dataset comprises a training set of 2,450 questions, a development set of 613 questions, and a test set of 751 questions. These questions cover three topics -- Ukrainian History, Ukrainian Language, and Ukrainian Literature -- and test both domain knowledge as well as reading comprehension.

\paragraph{Upper and Lower Sorbian MT} For DE-to-DSB MT, 171k translation pairs are provided as training data. In addition, a 4,000-pair development set is also provided for system validation and evaluation. Some monolingual sentences, approximately 10k, are also provided along with the translation data.

For DE-to-HSB MT, 187k training translation pairs and a 4,000-pair development set are provided. 300k monolingual sentences are further available for model enhancement. 

For both DSB and HSB, MCQ datasets are curated by the Witaj-Sprachzentrum. The questions are similar to the CEFR framework (A1 to C1), which follows the language certificate examinations. The difficulty of the questions ranges from simple true/false to complex multiple-choice formats with two to sixteen answer options. The development set contains 158 A1 to B2 level questions, while the test set for both languages has 205 questions for A1 to C1. An overview of the provided datasets is shown in Table \ref{tab:datasets}.

\begin{table*}[t]
\centering
\begin{tabularx}{\textwidth}{|p{2.5cm}|p{2cm}|p{2.3cm}|p{2.3cm}|X|}
\hline
\multicolumn{1}{|c|}{\textbf{Task}} &
\multicolumn{1}{c|}{\textbf{Training Data}} &
\multicolumn{1}{c|}{\textbf{Dev Set}} &
\multicolumn{1}{c|}{\textbf{Test Set}} &
\multicolumn{1}{c|}{\textbf{Notes}} \\ \hline\hline
EN$\rightarrow$UK Translation & None & 5,108 pairs & --- & Only dev set available \\ \hline
CS$\rightarrow$UK Translation & None & 6,263 pairs & --- & Only dev set available \\ \hline
Ukrainian QA (MCQs) & 2,450 questions & 613 questions & 751 questions & From ZNO exam; covers History, Language \& Literature; tests knowledge and comprehension \\ \hline
DE$\rightarrow$DSB Translation & 171k pairs & 4,000 pairs & --- & $\sim$10k monolingual sentences provided \\ \hline
DE$\rightarrow$HSB Translation & 187k pairs & 4,000 pairs & --- & $\sim$300k monolingual sentences provided \\ \hline
DSB QA (MCQs) & --- & 158 A1--B2 questions & 205 A1--C1 questions & From Witaj-Sprachzentrum; CEFR-style; difficulty from true/false to multiple-choice (2--16 options) \\ \hline
HSB QA (MCQs) & --- & 158 A1--B2 questions & 205 A1--C1 questions & Same as DSB QA \\ \hline
\end{tabularx}
\caption{Summary of the provided datasets for MT and QA.}
\label{tab:datasets}
\end{table*}

\subsection{Additional Data} \label{additiona_data}
\paragraph{Upper and Lower Sorbian MT} In order to enhance our DE--DSB/HSB translation systems, we incorporate additional parallel data. For DSB, we translate the provided 10k monolingual examples into German. Then we create 10k additional translation pairs using the translated German sentences and the monolingual DSB sentences. Since 10k is a small amount, we additionally use 24k DE--DSB sentences from the Tatoeba bilingual dataset.\footnote{\url{https://github.com/Helsinki-NLP/Tatoeba-Challenge/blob/master/data/README-v2023-09-26.md}} In turn, for HSB, we translate the first 100k monolingual sentences from the provided 300k sentences into German and create 100k additional translation pairs. Due to the substantial time required for translation, we translate only 100k sentences. To translate the DSB and HSB sentences into German, we first finetune two separate Qwen2.5-3B-Instruct models, one for HSB--DE and one for DSB--DE, using the respective provided parallel translation datasets. These models are finetuned by applying LoRA on all projection layers of the model. 

\paragraph{Upper Sorbian, Lower Sorbian, and Ukrainian QA} For QA in DSB, HSB, and Ukrainian, we select 16 English MCQA datasets, namely: (English) Global MMLU \cite{global_mmlu}, CommonsenseQA \cite{commonsenseqa}, ARC \cite{ai2arc}, Race \cite{race}, Dream \cite{dream}, PIQA \cite{piqa}, HellaSwag \cite{hellaswag}, SCIQ \cite{sciq}, MedMCQA \cite{medmcqa}, LogicQA \cite{logiqa}, Quail \cite{quail}, SocialIQA \cite{socialiqa}, CosmosQA \cite{cosmosqa}, OpenbookQA \cite{openbookqa}, QASC \cite{qasc}, BoolQ \cite{boolqa}. From these datasets, we sample up to 10k questions from each available split (training, development, and test), resulting in approximately 200k English MCQs. In order to translate the English MCQs into DSB and HSB, we first use \texttt{googletrans}\footnote{\url{https://github.com/ssut/py-googletrans}} to translate the German sentences of the provided DE--DSB and DE--HSB translation examples into English, in order to create English--DSB/HSB translation pairs. Second, using this  data, we finetune two separate Qwen2.5-3B-Instruct models on EN--DSB and EN--HSB MT. We use these two models to translate the English MCQs into DSB and HSB. For Ukrainian, we also translate the 200k English MCQs using \texttt{googletrans} directly, since Ukrainian is supported by Google Translate.

\paragraph{Ukrainian MT} For CS--UK MT, we collect training data from OpenSubtitles \cite{opus, opensub}, NeuTED \cite{neuted}, KDE4 \cite{opus}, and ELRC UKR Acts \cite{elrc}. For EN--UK, we use OpenSubtitles \cite{opus, opensub}, NeuTED \cite{neuted}, ELRC UKR Acts \cite{neuted}, and Multi30k \cite{multi30k_uk}. Across these sources, there are nearly 7 million sentence pairs per direction.

To reduce the number of sentence pairs, we apply a similarity-based retrieval method, using the provided development sets for  CS--UK and EN--UK as the reference datasets. We embed each Ukrainian sentence from this dataset by performing mean-pooling over the last hidden states of Qwen2.5-3B-Instruct token outputs. For each sentence of the reference dataset we retrieve the 75 most similar Ukrainian sentences from the collected pool of translation data along with the associated sentence in CS or EN and aggregate them. We then deduplicate the aggregated set to retain only unique translation pairs. Overall, we get 321k and 251k CS--UK and, respectively, EN--UK translation pairs for training. 
%We only consider 75 similar sentences for each sentence to create a compact yet smaller dataset, which we can use to train a model swiftly. 

\subsection{Data for Retrieval-augmented Generation} \label{rag}
For the Ukrainian QA task, we employ retrieval-augmented generation (RAG) using pages from Wikipedia and 10 books on Ukrainian history, language, and literature (the same sources used by the winning team of the UNLP 2024 Shared Task \cite{unlp}). We extract about 30k pages using the \texttt{wikipediaapi}\footnote{\url{https://github.com/martin-majlis/Wikipedia-API}} library, setting \texttt{max\_depth = 2} to include relevant subcategories from the Ukrainian history, language, and literature categories.

\section{Models and Algorithms}

\subsection{Qwen: The Underlying LLM}
All our models are LoRA \cite{lora}-finetuned Qwen2.5-3B-Instruct models, and, thus, satisfy the shared task’s parameter constraint. We use one model per language for all tasks.

\subsection{Finetuning}
\paragraph{Finetuning on MT and General QA Data}
For DSB and HSB, we combine the provided MT data, additional translations created by us, and translated MCQs to finetune Qwen2.5-3B-Instruct with LoRA \cite{lora} applied to all projection layers. For DSB/HSB, we use both the English and the translated version of each MCQ, in this format:

\begin{tcolorbox}[colback=gray!10, colframe=black, title=MCQ Prompt for DSB/HSB during training,
enhanced,
  breakable,
  sharp corners,
  boxrule=0.5pt,
  width=\linewidth,
  fontupper=\small]
\texttt{en\_context (if any)\\
dsb/hsb\_context (if any)\\\\
Question: \{en\_question\}\\
Question: \{dsb/hsb\_question\}\\\\
Possible Answers: \{en\_possible\_answers\}\\
Possible Answers: \{dsb/hsb\_possible\_ans\}\\\\
Answer: \{answer\}}
\end{tcolorbox}

%We insert exactly one space after \texttt{Answer:} before the correct option label (e.g., \texttt{Answer: C}), following \citet{sanzguerrero2025mindgapcloserlook}, who show that this spacing is essential for reliable parsing and evaluation.
To extract the model's predicted answer for QA, we end the prompt with ``\texttt{Answer:}'' and compute the next-token probabilities for each option label. The answer is then taken as the label with the highest probability. Following \citet{sanzguerrero2025mindgapcloserlook}, we evaluate the probabilities of the ``\texttt{\textvisiblespace X}'' tokens\footnote{Where ``\texttt{X}'' denotes one of the option labels.} (i.e., tokens formed by the preceding space \emph{together} with the option label), as this approach has been shown to yield better performance and calibration.

In order to use this prompt at inference time, we need to translate the provided DSB/HSB questions into English. For, first we finetuned two separate translation models, one for DSB--EN and another for HSB--EN, using the same dataset we use for EN--HSB and EN--HSB model finetuning, but this time in the opposite direction. These two translation models are also based on Qwen2.5-3B-Instruct  and trained with LoRA. During the development phase, we observe that for both DSB and HSB QA, alphabetic option labels lead to better results than numeric labels due to label bias \cite{zheng2024largelanguagemodelsrobust}. So, for training and evaluation, we use alphabetic option labels. 

The prompt for MT is of the following format:

\begin{tcolorbox}[colback=gray!10, colframe=black, title=MT Prompt during Training,
enhanced,
  breakable,
  sharp corners,
  boxrule=0.5pt,
  width=\linewidth,
  fontupper=\small]
\texttt{Translate this German sentence into Upper Sorbian. Put it in this format: <hsb> \{Upper Sorbian translation\} </hsb>\\<de> \{German Sentence\} </de>}
\end{tcolorbox}
For Lower Sorbian, ''Upper'' is replaced by ''Lower'' and <hsb> and </hsb> are replaced by <dsb> and </dsb>.   

For Ukrainian (UK), we train the model on MT and QA data described in Section \ref{additiona_data}.
 
We apply the default chat template for Qwen2.5 and train on complete instructions (system + user + assistant), as described in \citet{instruction_loss}. We use LoRA \cite{lora} for 10 epochs with an initial learning rate of $1e-4$ and a linear learning rate scheduler. We save the model checkpoint after every epoch. At the end of training, we select the model with the highest BLEU \cite{sacrebleu, ogbleu} score, i.e., we select the model using the MT development set.

\subsection{Final Finetuning on MT and In-domain QA Data for DSB and HSB}
After initial finetuning of the model, we conduct a second round of finetuning. This final model finetuning is also performed by applying LoRA to all projection layers. We follow the same procedure for DSB and HSB. The provided QA development sets for DSB and HSB contain a total of 158  questions each, from language difficulty levels A1-B2. In this second stage of finetuning, we use these 158 in-domain QA examples and the first 3k translation pairs used in the initial finetuning process. 
%We refer to these MCQ questions as in-domain because they follow the same distribution as the test set.
To mitigate data scarcity, we apply oversampling: each QA item is repeated five times. Then, we shuffle the MT and the oversampled QA set and finetune the first finetuned model again to improve domain alignment for QA. We add the 3k translation pairs to avoid catastrophic forgetting of the MT capability of the models.
The final models are used for both MT and QA during evaluation. As our dataset for the second round of finetuning is small (approximately 3.75k MT and QA examples), we tune the learning rate, searching over $1e-4, 1e-5, 1e-6$, and $1e-7$. Four separate models are trained with these learning rates exactly for one epoch. In this learning rate searching process, we exclude the questions from B2 during finetuning. We select the best learning rate based on performance for both QA (56 questions of B2 level) and MT (the first 400 samples of the 4k dev set). We follow this approach for both languages. After this experiment, we chose $1e-4$ for DSB and $1e-6$ for HSB. Then, we finetune the initial models with these learning rates for two epochs on approximately 3.75k instructions, including the questions from B2, performing no validation. 

\subsection{Averaging Probabilities}
During QA evaluation for DSB and HSB, we generate multiple responses for each question by shuffling the order of answer options. We perform this step to mitigate positional bias \cite{positional-bias} as much as possible. For questions with 2–3 options, we use all permutations, which are 2 and 6, respectively. For those with more than three options, we randomly sample 20 unique answer option orders. We compute the probability distribution over the answer options under the model for each order and average them; we select the option with the highest average likelihood as the final answer.

\subsection{Retrieval-augmented Generation for Ukrainian QA}
We segment the retrieved pages (see Section \ref{rag}) into chunks of 512 characters with an overlap of 64 characters and embed them using Qwen2.5-3B-Instruct. For each chunk, mean pooling is applied over the token representations obtained from the last hidden states. Embeddings are stored in two separate \texttt{ChromaDB} indexes: one for history and another for language and literature. We make embedding of each question following the same way we apply for page chunks, mean pooling over all token representations. Using the subject indicated for each question, we conduct a search in the corresponding subject-specific index. 
%This gives us strong signals due to the reduction of the search space and domain-specific chunks. 
At inference time, we retrieve the 5 most relevant chunks and use them as context alongside the question.

\subsection{Few-shot In-context Learning for MT}
For MT, we employ few-shot in-context learning using similarity-based retrieval, following \citet{mt-similarity}. For DSB/HSB, we embed each test-set German source sentence and retrieve the 5 most similar sentences from the development set, along with their translations. For Ukrainian, we use Ukrainian sentences for embedding generation and retrieval.

\begin{comment}
\subsection{Few-Shot Generation for QA}
For QA (DSB and HSB), we also employ a similarity search to retrieve a few-shot examples. First, we embed the concatenated context and question for each instance, then embed each instance. As the next step, we perform retrieval within the same CEFR language level (A1–B2) of the instance under consideration to ensure a similar difficulty level. 
\end{comment}

\section{Results and Discussion}
% Table 1: Translation - DSB & HSB
\begin{table}[t]
\centering
\resizebox{0.45\textwidth}{!}{%
\begin{tabular}{lcc}
\toprule
Model & \multicolumn{2}{c}{ChrF++} \\
\cmidrule(lr){2-3}
 & DSB & HSB \\
\midrule
Qwen2.5-3B-Instruct + LoRA(S2)(P) & 66.6 & 77.6 \\
Qwen2.5-3B-Instruct + LoRA(S1)    & 67.5 & 77.5 \\
\midrule
Baseline                          & 12.21 & 11.87 \\
\bottomrule
\end{tabular}%
}
\caption{ChrF++ scores for DSB and HSB. \textit{LoRA(S1)} = one round of LoRA finetuning; \textit{LoRA(S2)} = two rounds of LoRA finetuning. \textit{P} indicates our primary submission.}
\label{tab:results-both}
\end{table}

\paragraph{MT for Lower and Upper Sorbian}
Table~\ref{tab:results-both} shows that our proposed approach yields a significant improvement over the baseline. The baseline system achieves only 12.21 ChrF++ for DSB and 11.87 ChrF++ for HSB. The first round of LoRA finetuning, indicated by \textit{S1}, already increases ChrF++  to 67.5 and 77.5 for DE--DSB and DE--HSB, respectively. The goal for our second round of finetuning (\textit{S2}), is to adapt the model with in-domain QA data, while retaining the model's MT capability as much as possible. For HSB, \textit{S2}  slightly improves over S1 (77.6 vs. 77.5), but, for DSB, performance drops slightly, from 67.5 to 66.6. 

% Table 2: QA - DSB & HSB
\begin{table}[h]
\centering
\resizebox{0.45\textwidth}{!}{%
\begin{tabular}{lcc}
\toprule
Model & \multicolumn{2}{c}{Accuracy (\%)} \\
\cmidrule(lr){2-3}
 & DSB & HSB \\
\midrule
Qwen2.5-3B-Instruct + NO-FT           & 54.3 & 57.1 \\
Qwen2.5-3B-Instruct + LoRA(S1)        & 48.3 & 50.0 \\
Qwen2.5 + LoRA(S1) Avg.           & 50.7 & 54.3 \\
Qwen2.5-3B-Instruct + LoRA(S2)        & 48.3 & 50.5 \\
Qwen2.5 + LoRA(S2) Avg. (P)       & 51.7 & 55.2 \\
\midrule
Baseline                               & 45.85 & 42.86 \\
\bottomrule
\end{tabular}%
}
\caption{QA accuracy scores (A1--C1) for DSB and HSB. \textit{LoRA(S1)} = one round of LoRA finetuning; \textit{LoRA(S2)} = two rounds of LoRA finetuning; \textit{Avg.} = average over multiple option orders; \textit{NO-FT} = no finetuning, i.e., direct use of Qwen2.5-3B-Instruct. \textit{P} indicates our primary submission.}
\label{tab:qa-accuracy}
\end{table}

\paragraph{QA for Lower and Upper Sorbian}
For DSB and HSB QA (Table~\ref{tab:qa-accuracy}), the baseline accuracies of 45.85\% (DSB) and 42.86\% (HSB) are surpassed by all finetuned variants. Interestingly, the non-finetuned Qwen2.5-3B-Instruct model outperforms the baseline substantially, particularly for HSB (+14.24 accuracy). However, LoRA finetuning (S1 and S2) slightly reduces overall accuracy compared to the non-finetuned model, likely due to the trade-off introduced by joint MT and QA finetuning. 
%To mitigate these interference issues and also the positional bias, we introduced our variant-averaging method, which increases robustness to answer order permutations.
Averaging over the results for different answer option orders improves accuracy after both rounds of finetuning (S1 and S2). It helps more after the second round of finetuning, reaching 51.7\% for DSB and 55.2\% for HSB. The improvements over non-averaged results demonstrate that this straightforward method is effective for low-resource QA.
%, where overfitting to specific answer orderings might degrade performance.

\begin{table}[t]
\centering
\resizebox{0.45\textwidth}{!}{%
\begin{tabular}{lcc}
\toprule
Model & \multicolumn{2}{c}{ChrF++} \\
\cmidrule(lr){2-3}
 & CS--UK & EN--UK \\
\midrule
Qwen2.5-3B-Instruct + LoRA  & 8.09 & 3.10 \\
\midrule
Baseline & 3.48 & 0.40 \\
\bottomrule
\end{tabular}%
}
\caption{ChrF++ scores for CS--UK and EN--UK MT.}
\label{tab:csuk-enuk-chrf}
\end{table}

\paragraph{MT for Ukrainian}
The Ukrainian MT tasks (Table~\ref{tab:csuk-enuk-chrf}) are challenging due to the lack of  training data provided by the shared task. The baseline ChrF++ scores of 3.48 (CS--UK) and 0.40 (EN--UK) reflect the difficulty level. By retrieving and curating translation data via similarity search and then finetuning a Qwen2.5-3B-Instruct model, our system achieves slight improvements over the baseline for both CS--UK (8.09) and EN--UK (3.10). %These results indicate that our retrieval-based data curation can enhance performance. \
Though performance increases, the improvement is not as big as for the DE--DSB/HSB MT tasks.
%, even after using similarity-based few-shot learning. 
A possible reason for this is a mismatch between the training, development, and test sets: we train our models on sentences, but the test set consists of large documents and lengthy conversations.

\begin{table}[t]
\centering
\resizebox{0.45\textwidth}{!}{%
\begin{tabular}{lc}
\toprule
Model & Accuracy (\%) \\
\midrule
Qwen2.5-3B-Instruct + LoRA + RAG  & 35.82 \\
\midrule
Baseline                        & 31.16 \\
\bottomrule
\end{tabular}%
}
\caption{Accuracy scores for Ukrainian QA. The shown model is our primary submission.}
\label{tab:qa-uk}
\end{table}

\paragraph{QA for Ukrainian}
For Ukrainian QA (Table~\ref{tab:qa-uk}), our proposed model, based on finetuning jointly on MT and QA in combination with RAG, improves over the baseline: 31.16\% vs. 35.82\%. This gain is smaller than for DSB and HSB QA, which we attribute to two factors: First, the QA dataset for Ukrainian requires some factual knowledge regarding the Ukrainian language, history, and literature, which makes the task harder. Second, 
%and more critically for RAG, 
most LLMs are underexposed to the Cyrillic script, resulting in weaker tokenization, over-splitting of words, and a decreased quality of token representations \cite{unlp}. This results in poor-quality embeddings for Ukrainian sentences. As a result, retrieval quality degrades: semantically close passages are missed or under-ranked, and the injected context is less helpful. 

\section{Conclusion}
In this paper, we present JGU Mainz's submission to the WMT25 Shared Task on LLMs with Limited Resources for Slavic Languages, addressing MT and QA in Ukrainian, Upper Sorbian, and Lower Sorbian. Our approach combines parameter-efficient finetuning of Qwen2.5-3B-Instruct with training data augmentation and RAG for Ukrainian QA. %The MT \& QA joint finetuning pipeline 
Our primary submissions outperform the provided baselines for all languages and tasks, achieving substantial ChrF++ gains for DE–-DSB and DE-–HSB MT, as well as slight improvements for CS–-UK and EN-–UK MT. For QA, averaging over order options increases accuracy for both DSB and HSB, while, for Ukrainian, we achieve moderate gains through RAG.
%; however, we observed its impact varied across subject areas, highlighting the challenge of domain-specific questions. 
%Despite these improvements, several challenges remain: Ukrainian MT performance is hindered by domain mismatch between training and test data, since the training and development sets consist of sentences, but the test set consists of documents and long conversations; QA performance degrading due to task interference in joint MT and QA finetuning; and for both DSB/HSB QA, few-shot prompting sometimes reduces accuracy, warranting the requirement for further investigation.

\section*{Acknowledgments}
This work was supported by the Carl Zeiss Foundation through the TOPML and MAINCE projects (grant numbers P2021-02-014 and P2022-08-009I ).

\bibliography{custom}
\end{document}